\documentclass[10pt,twocolumn,letterpaper]{article}
\usepackage[T1]{fontenc}
\ifdefined\pdfminorversion
\fi
\PassOptionsToPackage{table}{xcolor}
\usepackage[pagenumbers]{cvpr}

\usepackage{makecell}
\usepackage{array}
\usepackage{tabularx}
\definecolor{cvprblue}{rgb}{0.21,0.49,0.74}
\usepackage[breaklinks,colorlinks,allcolors=cvprblue]{hyperref}
\hypersetup{
  pdftitle={ReTTT-AD: Test-Time Training Neighborhood Prediction for Zero-Shot Anomaly Detection},
  pdfauthor={Ke Xu, Peiyuan Li, Yanning Hou, Minghan Fan, Jianfeng Qiu}
}

\title{ReTTT-AD: Test-Time Training Neighborhood Prediction for Zero-Shot Anomaly Detection}
\author{
Ke Xu$^{1}$ \quad Peiyuan Li$^{1}$ \quad Yanning Hou$^{2,*}$ \\
Minghan Fan$^{1}$ \quad Jianfeng Qiu$^{1,*}$ \\[0.5em]
$^{1}$ Anhui University, Hefei, China \\
$^{2}$ National University of Defense Technology, Changsha, China
}

\begin{document}
\maketitle
\begingroup
\renewcommand{\thefootnote}{\fnsymbol{footnote}}
\footnotetext[1]{Corresponding authors.}
\endgroup

\begin{abstract}
Zero-shot anomaly detection (ZSAD) aims to identify anomalies in unseen categories without target-category training samples. Dominant methods learn normal and anomalous prototypes on an auxiliary dataset and detect anomalies via feature matching. However, this prototype-matching paradigm faces two key limitations: (i) heavy reliance on seen anomaly priors, which restricts generalization under unknown anomalies or domain shifts; and (ii) the neglect of intra-image local contextual relationships. To move beyond this paradigm, we conceptualize anomalies as spatial inconsistencies violating local contextual patterns and propose \textbf{ReTTT-AD}, establishing image-adaptive neighborhood prediction as the basis for anomaly discrimination. Specifically, our core Residual Local-TTT Block (RL-TTT) adapts a spatially shared predictor via a single-step Test-Time Training (TTT) update using neighborhood-to-center signals from the test image itself, modeling image-specific local regularities and capturing anomalies through prediction residuals. Furthermore, Normal-Tail Regularization (NTR) suppresses high-residual responses in complex normal regions, while a Global Anomaly Readout (GAR) module provides complementary image-level evidence. Extensive experiments across 14 benchmarks demonstrate that ReTTT-AD significantly outperforms state-of-the-art methods on both industrial and medical anomaly localization tasks.
\end{abstract}

\section{Introduction}
\label{sec:intro}

\begin{figure}[!htb]
    \centering 
    \includegraphics[width=0.47\textwidth]{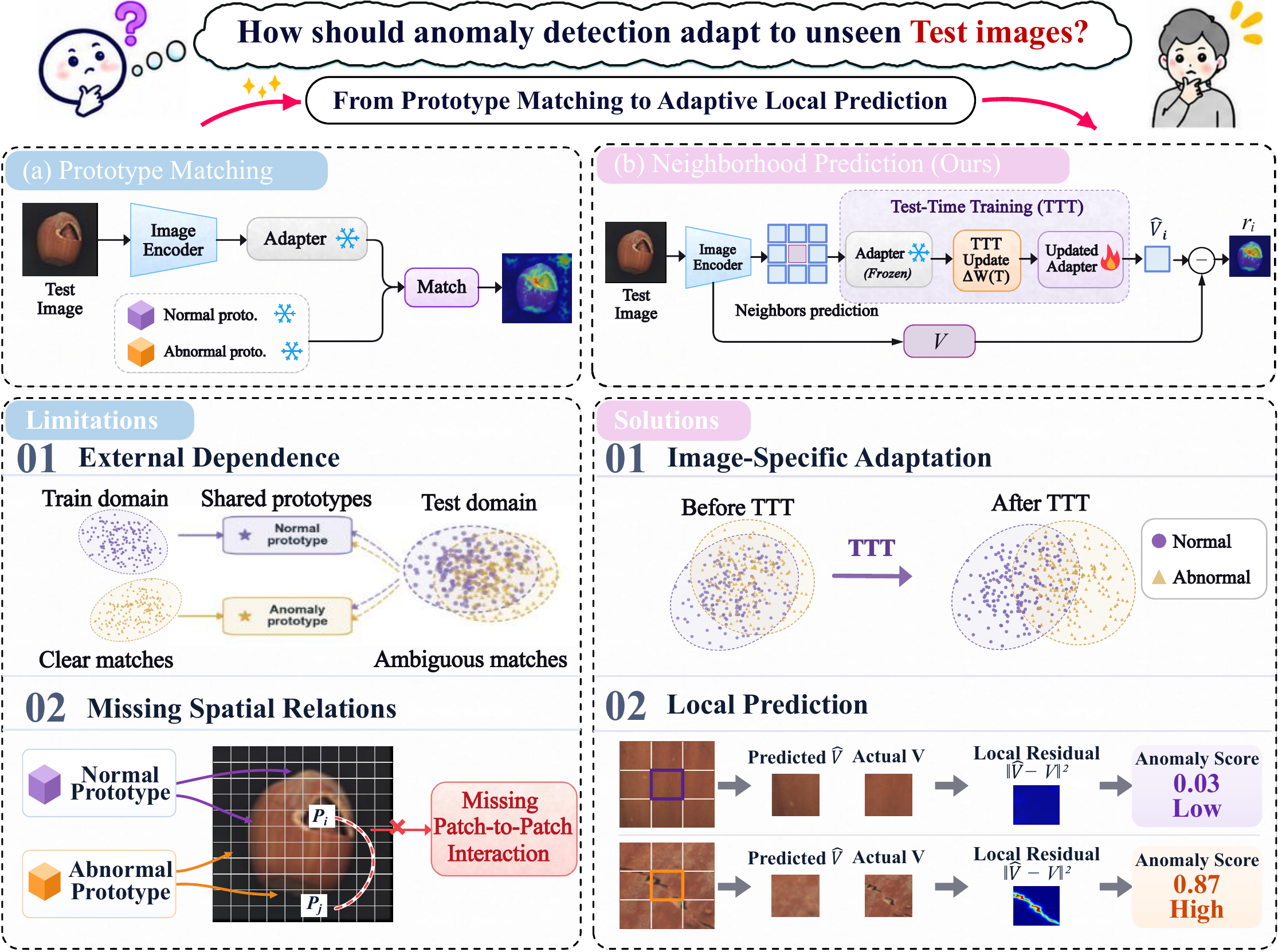} 
    \caption{\textbf{From static prototype matching to adaptive local prediction.}
    (a) Prototype matching relies on external normal and anomalous references, suffering from ambiguous matches under domain shifts and missing intra-image spatial relationships.
    (b) ReTTT-AD performs neighborhood-to-center prediction and adapts a shared predictor to each test image via TTT, using prediction residuals for local anomaly discrimination.}

    \label{fig:Intro}
\end{figure}

Anomaly detection aims to identify deviations from normal patterns in applications such as industrial inspection~\cite{MVTec,visa} and medical imaging~\cite{MFAD}. Traditional methods typically require target-category normal samples or anomaly annotations~\cite{PaDiM,PatchCore,KSDD2}, whose acquisition is costly for newly encountered categories. Zero-shot anomaly detection (ZSAD) addresses this limitation by generalizing to unseen categories and datasets without target-class training samples~\cite{WinCLIP,AnomalyCLIP,VisualAD}. Recent advances in pretrained vision and vision--language models have further facilitated this task through transferable visual representations~\cite{CLIP,DINOv2}.

Existing zero-shot anomaly detection (ZSAD) methods primarily rely on feature-to-prototype matching, utilizing text prompts or learned visual-semantic prototypes for anomaly discrimination~\cite{WinCLIP,AnomalyCLIP,VisualAD}. To enhance prototype adaptability, recent studies attempt to inject test-image global information into prototypes or prompt embeddings during inference~\cite{AdaCLIP,VCP,AdaptCLIP}. However, prototype-based methods suffer from two inherent limitations: (i) they heavily rely on seen anomaly priors from auxiliary datasets, limiting generalization under unknown anomalies or domain shifts; and (ii) globally conditioned prototypes only reflect dominant normal appearances, causing local anomalies to still maintain high similarity to normal prototypes and be misclassified as normal, even when their appearance remains nearly normal while their structural relationships with surrounding patches are disrupted. To overcome these limitations, we conceptualize anomalies as spatial inconsistencies violating local contextual patterns, and formulate local anomaly discrimination as an intra-image prediction task: predicting center-patch features from surrounding representations and measuring their prediction residuals. Because local structural relationships vary across test images, a fixed predictor may not provide a reliable reference for every image. We therefore introduce Test-Time Training (TTT)~\cite{sun2020test,gandelsman2022test} to adapt a spatially shared lightweight predictor through a single update derived solely from neighborhood-to-center signals in the current test image. This image-conditioned update induces location-dependent prediction corrections, allowing normal and anomalous regions to exhibit different residual responses for reliable anomaly discrimination.

However, local prediction residuals are imperfect anomaly indicators: complex normal textures may produce high residuals, while locally coherent anomalies may remain predictable. To address these complementary failure modes, we introduce NTR to suppress high-residual responses in normal regions and GAR to provide complementary global evidence for anomalies missed by local prediction. Building on these designs, we propose \textbf{ReTTT-AD}, a unified framework for zero-shot anomaly detection. It integrates image-conditioned local prediction with complementary global evidence, aggregating adapted residuals across layers and scales for pixel-level localization while combining local and global cues for image-level detection. 

In summary, our main contributions are as follows:
\begin{itemize}
    \item We reformulate local anomaly discrimination as image-conditioned neighborhood prediction and propose the Residual Local-TTT Block (RL-TTT), which adapts a spatially shared predictor to each test image and detects anomalies through prediction residuals. Controlled ablations and prediction-error analyses further examine the role of test-time adaptation.

    \item We introduce Normal-Tail Regularization (NTR) to suppress high-residual normal regions, improving the reliability of residual-based anomaly localization.

    \item We develop the Global Anomaly Readout (GAR), which leverages multi-layer CLS features for complementary image-level anomaly detection.

    \item Extensive experiments on 14 industrial and medical benchmarks demonstrate the effectiveness and cross-domain generalization of ReTTT-AD.
\end{itemize}

\section{Related Work}

\subsection{Zero-shot Anomaly Detection}
Zero-shot anomaly detection (ZSAD) aims to identify anomalies in unseen categories without target-class training samples. Relying on pretrained models like CLIP~\cite{CLIP}, existing methods primarily derive anomaly evidence from text prompts or learned visual-semantic prototypes. Specifically, WinCLIP~\cite{WinCLIP} and AnomalyCLIP~\cite{AnomalyCLIP} utilize multi-granularity matching and object-agnostic prompts. To enhance adaptability, AdaCLIP~\cite{AdaCLIP}, VCP-CLIP~\cite{VCP}, and FAPrompt~\cite{FAPrompt} dynamically refine prompt embeddings using visual context~\cite{tpt}, whereas AdaptCLIP~\cite{AdaptCLIP} adapts both visual and textual representations. Recently, VisualAD~\cite{VisualAD} explores language-free discrimination in the visual domain. Despite their effectiveness, they remain confined to feature-to-prototype matching with static predictor weights. In contrast, our approach exploits local structural relationships within the test image to establish explicit prediction-based anomaly evidence.

\subsection{Local Prediction and Test-Time Training}
Local feature modeling is widely used in anomaly detection~\cite{liu2023simplenet,CutPaste,gudovskiy2022cflow}. PaDiM~\cite{PaDiM} and PatchCore~\cite{PatchCore} rely on normal feature distributions and memory matching, while reconstruction- and prediction-based methods~\cite{Auto2,selfsupervised5} measure discrepancies between estimated and observed representations~\cite{deng2022anomaly}. Representative approaches further explore discriminative reconstruction, neighbor-masked feature modeling, and feature prediction with global reasoning~\cite{DRAEM,UniAD,EfficientAD}. However, their reliance on target-class normal data limits direct zero-shot applicability. Meanwhile, Test-Time Training (TTT) adapts models via self-supervised objectives during inference~\cite{liu2021ttt++}. Inspired by TTT, we formulate local feature modeling as an implicit neighborhood-to-center mapping and adapt a lightweight, spatially shared predictor to each test image while keeping the pretrained backbone frozen. The adapted predictor establishes an image-specific structural baseline, with post-update residuals serving as anomaly evidence.

\section{Approach}
\label{sec:approach}

\begin{figure*}[t]
    \centering
    \includegraphics[width=1\textwidth]{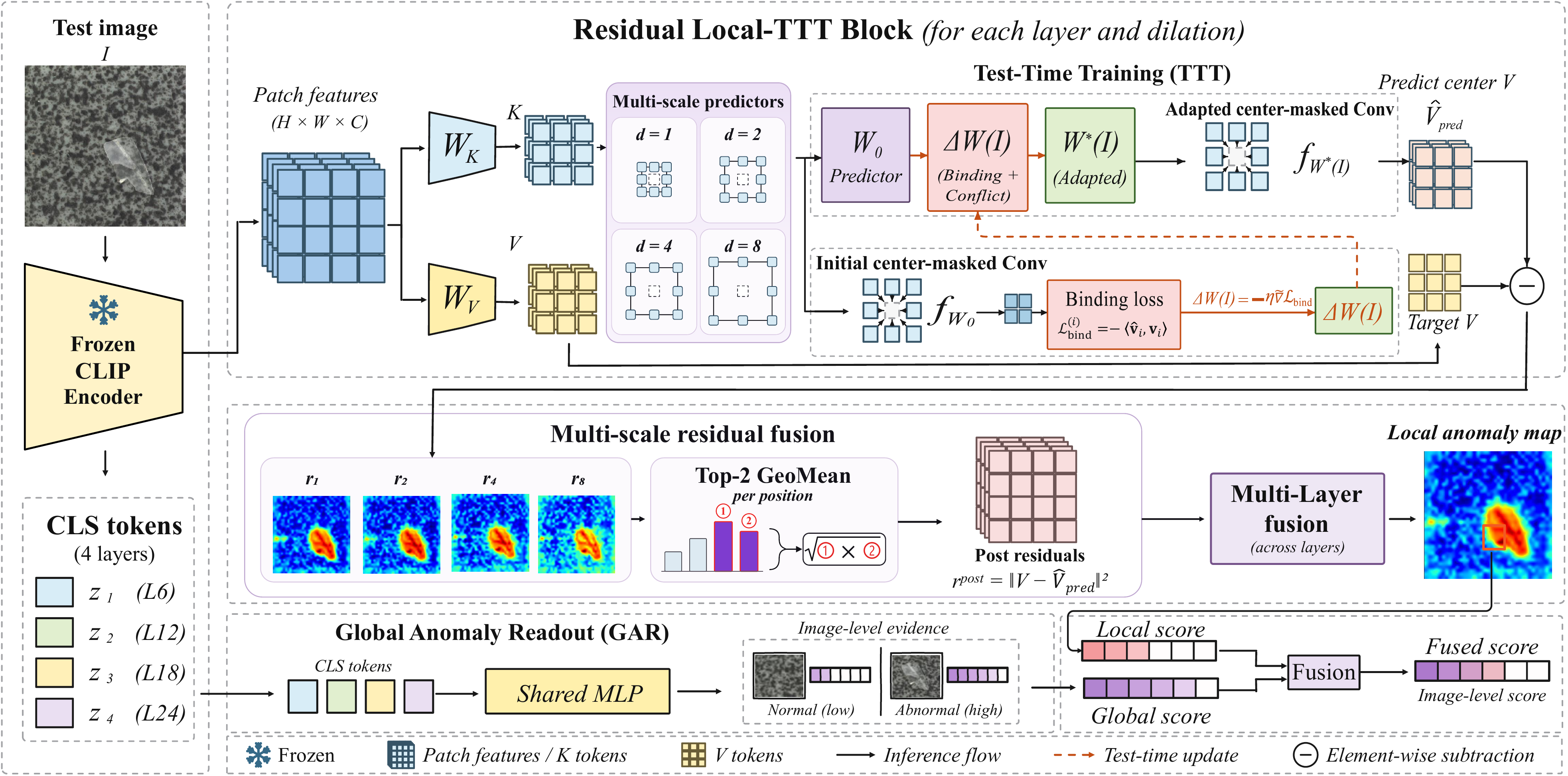}
    \caption{\textbf{Overview of ReTTT-AD.} A frozen visual encoder extracts multi-layer patch features and CLS representations. RL-TTT constructs center-masked neighborhood predictions and applies an image-conditioned update to a spatially shared predictor. The post-update prediction residuals are aggregated across layers and scales to generate anomaly maps. During source-domain training, NTR constrains high-residual normal patches to improve local discrimination, while GAR learns complementary image-level evidence from CLS features. At inference, local residual-based and global readout scores are combined for image-level anomaly detection.}
    \label{fig:Framework}
\end{figure*}

\subsection{Problem Setting and Overall Framework}
\label{sec:setting_framework}
We investigate zero-shot anomaly detection (ZSAD), where a model trained on source domains with pixel-level annotations is directly evaluated on unseen target categories without any fine-tuning for improved anomaly discrimination. 

To involve the test image in learning local prediction rules, our framework comprises three components: RL-TTT performs image-specific neighborhood prediction, NTR suppresses high residuals in normal regions, and GAR provides
complementary global evidence.

Given an image $I$, a frozen CLIP encoder extracts multi-layer patch features $\mathbf{X}^{(\ell)}\in\mathbb{R}^{N\times D}$ and CLS features $\mathbf{z}^{(\ell)}\in\mathbb{R}^D$ ($\ell\in\mathcal{L}$), where $N$ is the number of patches, $D$ is the backbone feature dimension, and $\mathcal{L}$ is the set of selected layers. RL-TTT processes patch features, while GAR processes CLS features. During testing, the relation predictor is only temporarily updated starting from the learned initial parameters, and updates do not accumulate across images.

\subsection{Residual Local-TTT Block (RL-TTT)}
\label{sec:rl_ttt}
To dynamically adapt to specific textures, RL-TTT adapts a lightweight neighborhood predictor through a single image-conditioned update step for each test image, followed by residual-based anomaly scoring. We omit layer and scale indices below for brevity.

\vspace{0.5em}
\noindent\textbf{Prediction Input and Target Construction.} 
We map the patch features into two representations:
\begin{equation}
    \mathbf{K} = \mathbf{X}\mathbf{W}_{\mathrm{K}}, \qquad \mathbf{V} = \mathbf{X}\mathbf{W}_{\mathrm{V}},
    \label{eq:rl_projection}
\end{equation}
where $\mathbf{W}_{\mathrm{K}}, \mathbf{W}_{\mathrm{V}} \in \mathbb{R}^{D\times C}$ are learnable linear projections, and $C$ is the projected feature dimension. The matrices $\mathbf{K},\mathbf{V}\in\mathbb{R}^{N\times C}$ contain the patch vectors $\mathbf{k}_i,\mathbf{v}_i\in\mathbb{R}^{C}$, respectively. $\mathbf{K}$ provides the input for neighborhood prediction, and $\mathbf{V}$ serves as the target for the center position. These projections remain fixed during testing.

\vspace{0.5em}
\noindent\textbf{Center-Masked Neighborhood Prediction.} 
We employ a lightweight channel-wise convolutional predictor to estimate the center vector $\mathbf{v}_i$ from neighboring vectors $\mathbf{k}_{i+\delta}$. Its $c$-th output component is:
\begin{equation}
    \hat{v}_{i,c} = \sum_{\delta\in\Omega_d} w_{c,\delta}\,k_{i+\delta,c},
    \label{eq:rl_prediction}
\end{equation}
where $c$ indexes feature channels, $w_{c,\delta}$ is an entry of the predictor $\mathbf{W}$, and $\Omega_d$ denotes the 8-neighborhood excluding the center at scale $d$. This operation masks the center, forcing the prediction to rely entirely on contextual information. The predictor $\mathbf{W}$ is shared spatially across the image.

\vspace{0.5em}
\noindent\textbf{Image-Specific Update \& Conflict Coordination.}
Directly minimizing prediction errors at test time may also reduce anomalous residuals, weakening anomaly contrast. To mitigate this risk, we introduce a binding objective that promotes alignment between neighborhood predictions and their corresponding center targets:
\begin{equation}
    \mathcal{L}_{\mathrm{bind}} = -\frac{\alpha}{N} \sum_{i=1}^{N} \left\langle\hat{\mathbf{v}}_i,\mathbf{v}_i\right\rangle,
    \label{eq:rl_binding}
\end{equation}
where $\alpha>0$ scales the binding objective. This objective avoids explicitly optimizing the scoring residuals. It yields a closed-form, non-iterative spatial gradient vector $\mathbf{g}_{i,c}$ for each channel $c$ at position $i$:
\begin{equation}
    \mathbf{g}_{i,c} = \frac{\partial \mathcal{L}_{\mathrm{bind}, i}}{\partial \mathbf{w}_{c}} = -\alpha\,\mathbf{k}_{\mathcal{N}(i),c} v_{i,c}.
    \label{eq:rl_gradient}
\end{equation}
Aggregating the per-position gradients gives the gradient of each kernel coefficient:
\begin{equation}
    \frac{\partial\mathcal{L}_{\mathrm{bind}}}{\partial w_{c,\delta}}
    = -\frac{\alpha}{N}\sum_i k_{i+\delta,c}v_{i,c}.
\end{equation}
This image-specific cross-correlation signal is independent of the initial predictor weights $\mathbf{W}_0$.

Because spatial gradients can conflict when anomalous patterns heavily distort local features, we introduce a gradient conflict-coordination strategy to limit interference. For each position $i$ and channel $c$, let $\mathbf{h}_{i,c} = \sum_{j \neq i} \mathbf{g}_{j,c}$ denote the aggregated gradient of all other positions, representing the dominant structural reference. We project out the conflicting component of $\mathbf{g}_{i,c}$ if it opposes this dominant direction, and then normalize the aggregate to yield a stabilized image-level gradient $\bar{\mathbf{G}}_c$:
\begin{equation}
\begin{aligned}
    \widetilde{\mathbf{g}}_{i,c} &= \mathbf{g}_{i,c} - \frac{\min(\langle\mathbf{g}_{i,c}, \mathbf{h}_{i,c}\rangle, 0)}{\|\mathbf{h}_{i,c}\|_2^2 + \varepsilon} \mathbf{h}_{i,c}, \\
    \bar{\mathbf{G}}_c &= \frac{\sum_i \widetilde{\mathbf{g}}_{i,c}}{1 + \left\lVert\sum_i \widetilde{\mathbf{g}}_{i,c}\right\rVert_2},
\end{aligned}
\label{eq:rl_conflict}
\end{equation}
where $\varepsilon=10^{-8}$. The resulting image-conditioned gradient is applied once as an additive correction to the learned initialization:
\[
\mathbf W^*(I)=\mathbf W_0-\eta\bar{\mathbf G}(I).
\]
Since the gradient is independent of the predictor weights, repeated steps would accumulate the same update direction. We therefore adopt a single step for efficient image-specific adaptation.

\vspace{0.5em}
\noindent\textbf{Adapted Residuals and Differential Correction.} 
Let $\hat{\mathbf{v}}_i^*$ denote the center prediction obtained by substituting the adapted predictor $\mathbf{W}^*(I)$ into Eq.~\eqref{eq:rl_prediction}. The channel-averaged squared residual is:
\begin{equation}
    r_{i,d}^{(\ell)} = \frac{1}{C} \left\lVert\hat{\mathbf{v}}_i^* - \mathbf{v}_i\right\rVert_2^2.
    \label{eq:rl_residual}
\end{equation}
Although the image-conditioned update $\bar{\mathbf{G}}(I)$ is shared across locations, its induced corrections depend heavily on local neighborhood features. 
Our analyses show that adaptation reduces prediction residuals in normal regions, whereas anomalous regions retain or exhibit increased prediction errors. These differential responses enhance the residual contrast between normal and anomalous regions.

\vspace{0.5em}
\noindent\textbf{Multi-Scale Relation Fusion.} 
To cover diverse structural sizes, we compute predictions across multiple neighborhood scales. After learnable scale calibration, we geometrically fuse the two maximum residuals at each position:
\begin{equation}
    r_i^{(\ell)} = \sqrt{\bar{r}_{i,(1)}^{(\ell)} \bar{r}_{i,(2)}^{(\ell)}},
    \label{eq:rl_scale_fusion}
\end{equation}
where $\bar{r}_{i,(1)}^{(\ell)}$ and $\bar{r}_{i,(2)}^{(\ell)}$ are the largest and second-largest calibrated residuals. Rather than relying on naive spatial voting, this geometric fusion effectively suppresses isolated single-scale noise and amplifies consistent structural deviations across receptive fields.

\subsection{Normal-Tail Regularization (NTR)}
\label{sec:ntr}
While local residuals effectively expose structural inconsistencies, complex normal textures and boundaries often yield spuriously high prediction errors, triggering severe false positives. Rather than uniformly penalizing all normal patches\textemdash{}which wastes optimization capacity on easily predictable backgrounds\textemdash{}NTR explicitly targets the upper tail of the normal residual distribution. 

\begin{figure}[t]
    \centering
    \includegraphics[width=\linewidth]{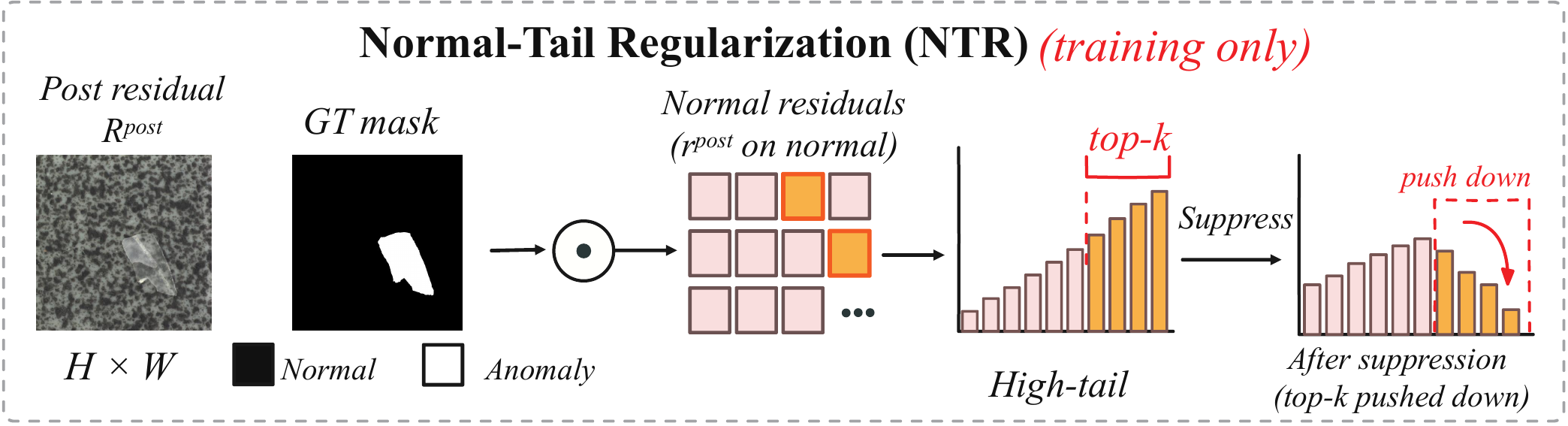}
    \caption{\textbf{Overview of Normal-Tail Regularization (NTR).}}
    \label{fig:ntr_module}
\end{figure}

During source-domain training, we dynamically select the top $\rho$ proportion of normal patches exhibiting the highest fused residuals (Eq.~\eqref{eq:rl_scale_fusion}), denoted as $\mathcal{T}_\rho^{(\ell)}$, and strictly minimize their average magnitude:
\begin{equation}
    \mathcal{L}_{\mathrm{NTR}} = \frac{1}{|\mathcal{L}|} \sum_{\ell\in\mathcal{L}} \frac{1}{|\mathcal{T}_\rho^{(\ell)}|} \sum_{i\in\mathcal{T}_\rho^{(\ell)}} r_i^{(\ell)}.
    \label{eq:rl_ntr_loss}
\end{equation}
By focusing optimization on hard-to-predict normal regions, NTR suppresses spurious high-residual responses and improves localization reliability without additional test-time overhead.

\subsection{Global Anomaly Readout (GAR)}
\label{sec:gar}
A fundamental vulnerability of purely local prediction is that locally coherent but globally anomalous structures (e.g., seamlessly repeating defective patterns) can be accurately predicted from their immediate surroundings, yielding deceptively low residuals. To seal this discrimination gap, we introduce the Global Anomaly Readout (GAR), deriving complementary image-level anomaly scores directly from the holistic CLS representations. 

Unlike the local pathway that evaluates neighborhood-to-center consistency, GAR provides an independent discrimination pathway anchored in full-image semantics:
\begin{equation}
    s_{\mathrm{GAR}}(I) = \frac{1}{|\mathcal{L}|} \sum_{\ell\in\mathcal{L}} f_\phi(\mathbf{z}^{(\ell)}).
    \label{eq:rl_gar_score}
\end{equation}
Supervised by the image-level label ($\mathcal{L}_{\mathrm{GAR}} = \operatorname{BCEWithLogits}(s_{\mathrm{GAR}},y)$), this module operates in parallel to the residual pathway. Fusing these two sources of evidence allows our framework to aggressively exploit fine-grained spatial discrepancies while improving sensitivity to anomalies insufficiently captured by local prediction throughout training and inference.

\subsection{Training Objective and Inference}
\label{sec:objective_inference}
During training, we compute the local anomaly logit map $\mathbf{A}^{(\ell)}$ by logarithmizing and upsampling $\mathbf{R}^{(\ell)}$. With $\mathbf{P}^{(\ell)} = \sigma(\mathbf{A}^{(\ell)})$ and the pixel-level ground-truth mask $\mathbf{Y}$, the segmentation loss is:
\begin{equation}
    \mathcal{L}_{\mathrm{seg}}
    = \sum_{\ell\in\mathcal{L}}
    \left(\mathcal{L}_{\mathrm{focal}}^{(\ell)}
    + \mathcal{L}_{\mathrm{dice}}^{(\ell)}\right).
    \label{eq:rl_seg_loss}
\end{equation}
Averaging the top-$k$ responses of $\sum_{\ell\in\mathcal{L}} \mathbf{A}^{(\ell)}$ yields the local image score, supervised by $\mathcal{L}_{\mathrm{local\text{-}img}}$. The total outer objective combines the segmentation loss in Eq.~\eqref{eq:rl_seg_loss}, NTR, and GAR supervision with the local image loss:
\begin{equation}
    \mathcal{L}_{\mathrm{outer}}
    = \mathcal{L}_{\mathrm{seg}}
    + \mathcal{L}_{\mathrm{local\text{-}img}}
    + \widetilde{\mathcal{L}}_{\mathrm{NTR}}
    + \widetilde{\mathcal{L}}_{\mathrm{GAR}},
    \label{eq:rl_outer_loss}
\end{equation}
where $\widetilde{\mathcal{L}}_{\mathrm{NTR}}$ and $\widetilde{\mathcal{L}}_{\mathrm{GAR}}$ denote the weighted NTR and GAR losses.

During inference, the pixel-level anomaly map is obtained by Gaussian smoothing ($\mathcal{G}$):
\begin{equation}
    \mathbf{A}(I) = \mathcal{G} \left( \sum_{\ell\in\mathcal{L}} \operatorname{Up} \bigl( \log \mathbf{R}^{(\ell)} \bigr) \right).
    \label{eq:rl_pixel_map}
\end{equation}
The final image-level prediction fuses the top-$k$ mean of $\mathbf{A}(I)$ (denoted as $s_{\mathrm{local}}$) with the global score:
\begin{equation}
    s_{\mathrm{img}} = (1-\lambda) s_{\mathrm{local}} + \lambda s_{\mathrm{GAR}},
    \label{eq:rl_image_score}
\end{equation}
where $k$ is the number of highest-scoring pixels retained for pooling and $\lambda\in[0,1]$ controls local--global score fusion.

\begin{table*}[!t]
\centering
\begingroup
\newcommand{\tblbest}[1]{\textcolor{red!75!black}{\textbf{#1}}}
\newcommand{\tblsecond}[1]{\textcolor{blue!75!black}{#1}}
\caption{Comparison on industrial benchmarks. Image-level metrics: (AUROC, $F_1$-max, AP); pixel-level metrics: (AUROC, $F_1$-max, AP, PRO), all in percent. $\textcolor{red}{\boldsymbol{\checkmark}}$/$\textcolor{green!70!black}{\boldsymbol{\times}}$ denote language-free/not language-free. $\dagger$ and $\ddagger$ indicate CLIP and DINOv2 backbones. Best and second-best values are red and blue.}
\label{tab:industrial_comparison}
\begingroup
\scriptsize
\setlength{\tabcolsep}{2pt}
\renewcommand{\arraystretch}{1.16}
\resizebox{\textwidth}{!}{%
\begin{tabular}{cl*{5}{c}*{2}{>{\columncolor{gray!12}}c}}
\toprule
\textbf{Metric} & \textbf{Dataset} & AnomalyCLIP & AdaCLIP & FAPrompt & AdaptCLIP & VisualAD$^{\dagger}$ & \textbf{Ours}$^{\dagger}$ & \textbf{Ours}$^{\ddagger}$ \\
 &  & $\textcolor{green!70!black}{\boldsymbol{\times}}$ & $\textcolor{green!70!black}{\boldsymbol{\times}}$ & $\textcolor{green!70!black}{\boldsymbol{\times}}$ & $\textcolor{green!70!black}{\boldsymbol{\times}}$ & $\textcolor{red}{\boldsymbol{\checkmark}}$ & $\textcolor{red}{\boldsymbol{\checkmark}}$ & $\textcolor{red}{\boldsymbol{\checkmark}}$ \\
\midrule
 & MVTec-AD & (91.6,92.7,96.2) & (92.0,92.7,96.4) & (91.8,92.3,95.6) & (92.5,93.2,96.4) & (92.2,93.2,96.7) & (\tblsecond{93.5},\tblsecond{93.3},\tblsecond{97.1}) & (\tblbest{93.9},\tblbest{94.5},\tblbest{97.5}) \\
 & VisA & (81.0,80.3,84.4) & (79.7,79.6,83.2) & (85.0,82.6,86.7) & (85.2,82.8,88.0) & (84.7,82.5,87.6) & (\tblsecond{88.4},\tblsecond{84.5},\tblsecond{90.4}) & (\tblbest{89.0},\tblbest{85.2},\tblbest{90.6}) \\
 & BTAD & (88.7,86.0,90.6) & (90.0,87.2,91.5) & (90.6,82.3,87.1) & (93.3,87.8,91.8) & (\tblsecond{94.9},\tblsecond{93.9},\tblsecond{97.0}) & (\tblbest{96.2},\tblbest{95.0},\tblbest{98.5}) & (89.7,85.7,87.9) \\
\raisebox{-2ex}[0pt][0pt]{\rotatebox[origin=c]{90}{\makecell{\textbf{Image-Level}}}} & KSDD2 & (91.9,84.5,93.4) & (94.9,90.3,96.2) & (95.5,89.5,96.5) & (95.7,90.1,96.7) & (\tblbest{98.0},\tblbest{93.9},\tblbest{98.3}) & (97.3,\tblsecond{92.4},97.8) & (\tblsecond{97.6},92.0,\tblsecond{97.9}) \\
 & DAGM & (98.0,90.6,92.4) & (98.3,91.5,94.2) & (99.1,93.7,96.5) & (96.1,87.6,89.2) & (\tblsecond{99.5},95.0,\tblsecond{97.8}) & (\tblbest{99.6},\tblbest{96.2},\tblbest{98.2}) & (99.4,\tblsecond{95.1},97.7) \\
 & DTD-Synthetic & (93.7,94.3,97.4) & (92.1,92.4,96.3) & (95.6,94.3,97.9) & (93.5,93.1,97.1) & (\tblbest{97.5},\tblsecond{96.6},\tblbest{99.1}) & (\tblsecond{97.1},96.0,\tblsecond{98.9}) & (95.3,\tblbest{96.7},98.7) \\
 & MVTec2 & (57.4,80.6,73.8) & (60.0,80.1,75.6) & (63.7,80.7,\tblsecond{77.9}) & (57.4,80.5,74.6) & (\tblsecond{64.0},\tblsecond{81.0},77.0) & (61.6,80.5,75.9) & (\tblbest{69.5},\tblbest{81.9},\tblbest{79.2}) \\
\cmidrule(lr){2-9}
 & \textbf{Mean} & (86.0,87.0,89.7) & (86.7,87.7,90.5) & (88.7,87.9,91.2) & (87.7,87.9,90.5) & (90.1,\tblsecond{90.9},\tblsecond{93.4}) & (\tblsecond{90.5},\tblbest{91.1},\tblbest{93.8}) & (\tblbest{90.6},90.2,92.8) \\
\midrule
 & MVTec-AD & (91.0,38.9,34.4,81.7) & (88.5,43.9,41.0,47.6) & (89.4,40.8,36.9,85.1) & (90.6,42.9,37.8,83.2) & (90.8,43.9,41.2,87.5) & (\tblbest{91.4},\tblsecond{45.7},\tblsecond{44.2},\tblbest{88.5}) & (\tblsecond{91.3},\tblbest{46.5},\tblbest{44.6},\tblsecond{88.3}) \\
 & VisA & (95.4,27.6,20.7,86.4) & (95.1,33.8,\tblsecond{29.2},71.3) & (\tblbest{95.8},28.8,22.6,88.6) & (95.5,32.5,25.7,84.7) & (\tblbest{95.8},34.6,28.4,\tblsecond{91.0}) & (\tblsecond{95.5},\tblsecond{34.7},28.3,\tblbest{91.4}) & (95.4,\tblbest{34.9},\tblbest{30.0},90.8) \\
 & BTAD & (93.0,47.1,41.5,71.0) & (87.7,42.3,36.6,17.1) & (92.3,46.3,39.1,74.4) & (96.4,47.8,44.5,78.2) & (91.1,49.8,43.1,\tblsecond{80.4}) & (\tblbest{94.5},\tblbest{53.5},\tblbest{49.8},\tblbest{84.1}) & (\tblsecond{93.8},\tblsecond{50.5},\tblsecond{45.5},79.1) \\
\raisebox{-4ex}[0pt][0pt]{\rotatebox[origin=c]{90}{\makecell{\textbf{Pixel-Level}}}} & KSDD2 & (98.0,50.6,43.7,90.8) & (96.1,59.2,58.6,40.8) & (\tblsecond{99.3},55.2,50.7,96.9) & (99.3,59.5,54.8,96.0) & (98.5,60.9,62.1,98.5) & (\tblbest{99.6},\tblbest{62.3},\tblbest{65.3},\tblsecond{98.9}) & (\tblbest{99.6},\tblsecond{61.0},\tblsecond{62.8},\tblbest{99.3}) \\
 & DAGM & (\tblsecond{96.9},56.9,53.6,89.2) & (88.6,48.5,42.6,37.6) & (\tblbest{98.1},56.5,54.4,\tblbest{95.3}) & (95.5,39.3,37.1,82.5) & (92.2,57.9,56.4,89.3) & (96.2,\tblsecond{59.2},\tblsecond{59.3},\tblsecond{92.7}) & (94.6,\tblbest{61.2},\tblbest{61.2},91.4) \\
 & DTD-Synthetic & (97.5,55.8,52.5,87.9) & (95.1,58.4,56.1,34.3) & (97.7,57.5,55.9,92.0) & (97.6,58.7,60.1,87.7) & (\tblsecond{98.1},64.3,65.5,\tblsecond{94.8}) & (97.6,\tblsecond{64.6},\tblsecond{66.5},94.4) & (\tblbest{98.5},\tblbest{66.0},\tblbest{69.5},\tblbest{95.3}) \\
 & MVTec2 & (81.9,16.5,10.3,41.5) & (76.5,17.8,\tblsecond{14.1},30.3) & (\tblsecond{82.0},18.4,12.2,45.8) & (80.3,17.4,12.5,48.5) & (79.6,17.9,13.3,\tblsecond{59.7}) & (81.6,\tblbest{19.4},\tblbest{14.5},55.4) & (\tblbest{84.0},\tblsecond{19.0},14.0,\tblbest{62.3}) \\
\cmidrule(lr){2-9}
 & \textbf{Mean} & (93.4,41.9,36.7,78.4) & (89.7,43.4,39.7,39.9) & (93.5,43.4,38.8,82.6) & (93.6,42.6,38.9,80.1) & (92.3,47.0,\tblsecond{44.3},85.9) & (\tblsecond{93.8},\tblbest{48.5},\tblbest{46.8},\tblsecond{86.5}) & (\tblbest{93.9},\tblsecond{48.4},\tblbest{46.8},\tblbest{86.6}) \\
\bottomrule
\end{tabular}%
}
\endgroup

\par\vspace{1pt}
\setlength{\abovecaptionskip}{2pt}
\setlength{\belowcaptionskip}{2pt}
\caption{Comparison on medical benchmarks. Notation follows Table~\ref{tab:industrial_comparison}.}
\label{tab:medical_comparison}
\begingroup
\scriptsize
\setlength{\tabcolsep}{2pt}
\renewcommand{\arraystretch}{1.16}
\resizebox{\textwidth}{!}{%
\begin{tabular}{cl*{5}{c}*{2}{>{\columncolor{gray!12}}c}}
\toprule
\textbf{Metric} & \textbf{Dataset} & AnomalyCLIP & AdaCLIP & FAPrompt & AdaptCLIP & VisualAD$^{\dagger}$ & \textbf{Ours}$^{\dagger}$ & \textbf{Ours}$^{\ddagger}$ \\
 &  & $\textcolor{green!70!black}{\boldsymbol{\times}}$ & $\textcolor{green!70!black}{\boldsymbol{\times}}$ & $\textcolor{green!70!black}{\boldsymbol{\times}}$ & $\textcolor{green!70!black}{\boldsymbol{\times}}$ & $\textcolor{red}{\boldsymbol{\checkmark}}$ & $\textcolor{red}{\boldsymbol{\checkmark}}$ & $\textcolor{red}{\boldsymbol{\checkmark}}$ \\
\midrule
 & OCT17 & (63.7,85.7,86.5) & (77.3,86.1,91.8) & (67.9,85.7,87.0) & (70.5,85.7,89.5) & (\tblsecond{88.9},88.2,\tblsecond{96.4}) & (\tblsecond{88.9},\tblsecond{89.1},96.3) & (\tblbest{94.8},\tblbest{93.1},\tblbest{98.2}) \\
\raisebox{-2ex}[0pt][0pt]{\rotatebox[origin=c]{90}{\makecell{\textbf{Image-Level}}}} & BrainMRI & (96.4,\tblsecond{94.7},97.2) & (94.9,93.1,96.8) & (95.9,\tblsecond{94.7},95.4) & (95.1,93.6,95.6) & (\tblsecond{96.7},\tblbest{94.8},\tblsecond{97.6}) & (\tblbest{97.2},93.9,\tblbest{98.0}) & (95.0,91.6,97.0) \\
 & HIS & (55.2,66.8,\tblsecond{56.1}) & (59.9,68.4,55.8) & (\tblbest{61.5},67.1,\tblbest{57.6}) & (61.9,67.8,57.9) & (60.1,\tblbest{68.9},55.9) & (\tblsecond{60.3},\tblsecond{68.6},55.4) & (\tblsecond{60.3},68.1,54.6) \\
\cmidrule(lr){2-9}
 & \textbf{Mean} & (71.8,82.4,79.9) & (77.4,82.5,81.5) & (75.1,82.5,80.0) & (75.8,82.4,81.0) & (81.9,\tblsecond{84.0},\tblbest{83.3}) & (\tblsecond{82.1},83.9,\tblsecond{83.2}) & (\tblbest{83.4},\tblbest{84.3},\tblbest{83.3}) \\
\midrule
 & ISIC & (88.5,\tblsecond{71.0},74.8,78.1) & (88.4,70.0,74.1,35.5) & (\tblsecond{89.2},70.8,\tblbest{77.1},\tblsecond{80.0}) & (89.3,70.9,77.3,76.6) & (84.0,65.1,67.7,73.2) & (86.4,68.7,72.1,77.6) & (\tblbest{90.0},\tblbest{73.4},\tblsecond{76.1},\tblbest{80.4}) \\
\raisebox{-4ex}[0pt][0pt]{\rotatebox[origin=c]{90}{\makecell{\textbf{Pixel-Level}}}} & CVC-ClinicDB & (84.6,45.5,41.7,69.8) & (84.3,44.4,43.3,63.6) & (83.5,46.5,\tblsecond{43.7},69.8) & (85.2,43.3,39.5,68.6) & (85.2,45.6,37.8,70.6) & (\tblbest{88.1},\tblbest{50.9},\tblbest{45.6},\tblbest{75.8}) & (\tblsecond{87.9},\tblsecond{49.0},42.6,\tblsecond{73.8}) \\
 & Endo & (86.5,53.7,50.7,67.1) & (82.9,47.9,47.7,55.8) & (86.4,55.1,\tblsecond{53.9},69.1) & (88.1,56.2,52.6,69.6) & (84.9,52.0,46.2,66.3) & (\tblsecond{87.4},\tblsecond{56.2},51.4,\tblsecond{71.8}) & (\tblbest{89.4},\tblbest{59.8},\tblbest{57.1},\tblbest{74.6}) \\
 & Kvasir & (82.0,50.0,43.2,44.7) & (80.3,47.7,31.0,27.5) & (81.1,49.0,44.8,44.7) & (86.0,55.3,48.0,39.5) & (80.3,47.8,47.6,\tblbest{50.8}) & (\tblsecond{83.6},\tblsecond{53.1},\tblsecond{47.9},\tblsecond{47.9}) & (\tblbest{87.0},\tblbest{57.8},\tblbest{54.1},40.4) \\
\cmidrule(lr){2-9}
 & \textbf{Mean} & (85.4,55.1,52.6,64.9) & (84.0,52.5,49.0,45.6) & (85.1,55.3,\tblsecond{54.9},65.9) & (87.2,56.4,54.4,63.6) & (83.6,52.6,49.8,65.2) & (\tblsecond{86.4},\tblsecond{57.2},54.3,\tblbest{68.3}) & (\tblbest{88.6},\tblbest{60.0},\tblbest{57.5},\tblsecond{67.3}) \\
\bottomrule
\end{tabular}%
}
\endgroup
\endgroup
\end{table*}

\section{Experiments}
\label{sec:experiments}

\subsection{Experimental Setup}
\label{sec:experimental_setup}

\noindent\textbf{Datasets.}
We evaluate our framework on 14 industrial and medical datasets. The seven industrial benchmarks are MVTec-AD~\cite{MVTec}, VisA~\cite{visa}, BTAD~\cite{BTAD}, KSDD2~\cite{KSDD2}, DAGM~\cite{DAGM}, DTD-Synthetic~\cite{DTD-Synthetic}, and MVTec2~\cite{MVTecAD2}. The seven medical benchmarks include OCT17~\cite{OCT17}, BrainMRI~\cite{BrainMRI}, and HIS~\cite{HIS} for image-level detection, and ISIC~\cite{ISIC}, CVC-ClinicDB~\cite{CVC-ClinicDB}, Endo~\cite{Endo}, and Kvasir~\cite{Kvasir} for pixel-level localization. We train on VisA and evaluate on the remaining datasets, using MVTec-AD as the source when VisA is the target. Target-domain labels are used only for evaluation, while TTT relies solely on the current test image for each target dataset.

\vspace{\baselineskip}
\noindent
\textbf{Evaluation metrics.}  
For classification, we report image-level metrics including the Area Under the ROC Curve (AUROC), the maximum $F_1$ score over all thresholds ($F_1$-max), and Average Precision (AP).
For segmentation, we evaluate pixel-level AUROC,  the maximum $F_1$ score over all thresholds ($F_1$-max), AP, and Per-Region Overlap (PRO) to assess the localization capability of the model.

\vspace{\baselineskip}
\noindent
\textbf{Implementation details.}
We employ frozen CLIP ViT-L/14@336px~\cite{CLIP} and DINOv2 ViT-L/14~\cite{DINOv2} backbones, resizing input images to $518\times518$ and extracting features from layers $\{6,12,18,24\}$. RL-TTT projects patch features into 64-dimensional $K$ and $V$ representations for local neighborhood prediction. A single binding-based update with conflict coordination adapts the lightweight predictor independently for each test image. The trainable modules are optimized using AdamW with a batch size of $8$ and a learning rate of $10^{-4}$. NTR regularizes high-residual normal patches, while GAR provides complementary global anomaly scores.

\subsection{Performance Comparison with SOTA Methods}
\label{sec:main_comparison}
We compare with AnomalyCLIP~\cite{AnomalyCLIP}, AdaCLIP~\cite{AdaCLIP}, FAPrompt~\cite{FAPrompt}, AdaptCLIP~\cite{AdaptCLIP}, and VisualAD~\cite{VisualAD}, and evaluate ReTTT-AD with a DINOv2 backbone. As shown in Tables~\ref{tab:industrial_comparison} and~\ref{tab:medical_comparison}, ReTTT-AD achieves competitive image-level classification and strong pixel-level localization across industrial and medical benchmarks. Under the identical CLIP backbone, ReTTT-AD consistently outperforms VisualAD, improving mean pixel AP from $44.3$ to $46.8$ across seven industrial datasets and from $49.8$ to $54.3$ across four medical datasets. Beyond these improvements, the CLIP variant achieves an image AUROC of $97.2$ on BrainMRI and a pixel AP of $45.6$ on CVC-ClinicDB, demonstrating its effectiveness in brain tumor detection and endoscopic lesion localization. Furthermore, with a frozen DINOv2 backbone, ReTTT-AD achieves strong pixel-level performance, with a mean pixel AP of $57.5$ across medical benchmarks and $76.1$ on ISIC. These results demonstrate that our relational prediction mechanism is compatible with different pretrained visual representations and generalizes across diverse imaging domains. Industrial source-domain training provides a transferable prediction foundation, while TTT adapts the predictor to image-specific neighborhood relationships without any target-domain supervision.

\subsection{Ablation Studies}
\label{sec:ablations}
We organize the ablations around three core questions: whether updating the local relational predictor effectively establishes an image-specific reference, whether normal-tail regularization (NTR) improves residual reliability, and whether global evidence (GAR) complements local spatial anomalies. We use a frozen CLIP ViT-L/14@336px and evaluate transfer on VisA$\rightarrow$MVTec and MVTec$\rightarrow$VisA.

\vspace{0.5em}
\noindent\textbf{Contribution of the Three Components.}
Table~\ref{tab:component_ablation} shows the complementary contributions of RL-TTT, NTR, and GAR. Removing RL-TTT substantially degrades pixel-level localization, confirming image-conditioned neighborhood prediction is essential for spatial discrimination. In this ablation, the local relation prediction branch is replaced with direct matching to normal and anomalous tokens. NTR suppresses high residuals in complex normal regions, while GAR provides complementary global supervision.

\begin{table}[!htbp]
    \centering
    \caption{Component ablation. Each level reports (AUROC, $F_1$, AP) in \%. Bold marks the best result per direction. Note: Without RL-TTT, the local relation-prediction branch is replaced with direct matching to normal and anomalous tokens.}
    \label{tab:component_ablation}
    
    \footnotesize 
    \renewcommand{\arraystretch}{1.15} 
    \setlength{\tabcolsep}{2.1mm}

    \begin{tabular}{c c c c c}
        \toprule
        \textbf{RL-TTT} & \textbf{NTR} & \textbf{GAR} 
        & \textbf{Pixel-level} 
        & \textbf{Image-level} \\
        \midrule

        \multicolumn{5}{c}{\textbf{VisA $\boldsymbol{\rightarrow}$ MVTec}} \\
        \midrule

        $\boldsymbol{\times}$ & $\boldsymbol{\checkmark}$ & $\boldsymbol{\checkmark}$ 
        & (82.7, 26.0, 18.3) & (85.0, 90.0, 92.6) \\

        $\boldsymbol{\checkmark}$ & $\boldsymbol{\times}$ & $\boldsymbol{\checkmark}$ 
        & (90.5, 44.3, 42.1) & (92.3, 93.0, 96.5) \\

        $\boldsymbol{\checkmark}$ & $\boldsymbol{\checkmark}$ & $\boldsymbol{\times}$ 
        & (91.4, 45.7, 44.2) & (90.6, 91.9, 95.8) \\

        \rowcolor{gray!12}
        $\boldsymbol{\checkmark}$ & $\boldsymbol{\checkmark}$ & $\boldsymbol{\checkmark}$ 
        & (\textbf{91.4}, \textbf{45.7}, \textbf{44.2}) & (\textbf{93.5}, \textbf{93.3}, \textbf{97.1}) \\

        \midrule
        \multicolumn{5}{c}{\textbf{MVTec $\boldsymbol{\rightarrow}$ VisA}} \\
        \midrule

        $\boldsymbol{\times}$ & $\boldsymbol{\checkmark}$ & $\boldsymbol{\checkmark}$ 
        & (92.5, 21.3, 14.7) & (80.5, 79.3, 83.8) \\

        $\boldsymbol{\checkmark}$ & $\boldsymbol{\times}$ & $\boldsymbol{\checkmark}$ 
        & (95.1, 31.9, 24.7) & (85.2, 82.1, 87.9) \\

        $\boldsymbol{\checkmark}$ & $\boldsymbol{\checkmark}$ & $\boldsymbol{\times}$ 
        & (95.5, 34.7, 28.3) & (85.5, 82.6, 88.2) \\

        \rowcolor{gray!12}
        $\boldsymbol{\checkmark}$ & $\boldsymbol{\checkmark}$ & $\boldsymbol{\checkmark}$ 
        & (\textbf{95.5}, \textbf{34.7}, \textbf{28.3}) & (\textbf{88.4}, \textbf{84.5}, \textbf{90.4}) \\

        \bottomrule
    \end{tabular}
\end{table}

\begin{figure*}[t]
  \centering
  \includegraphics[width=\textwidth]{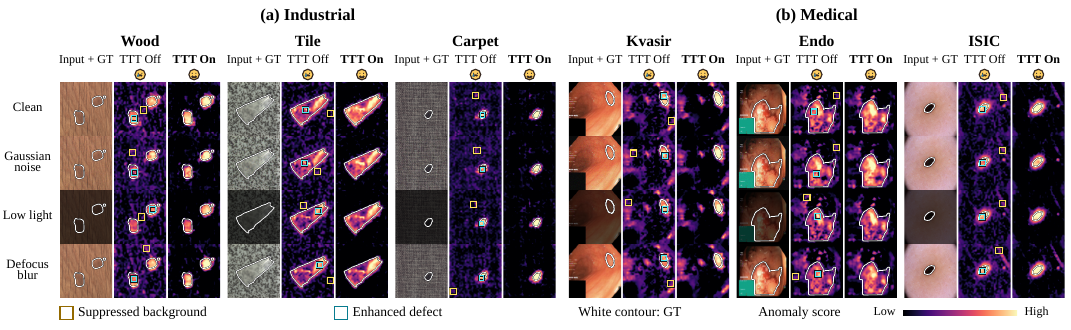}
  \caption{\textbf{Test-Time Training (\textit{TTT On}) enhances normal-anomalous residual contrast across domains and visual corruptions.} Qualitative comparisons on (a) industrial (Wood, Tile, Carpet) and (b) medical (Kvasir, Endo, ISIC) benchmarks under clean, Gaussian noise, low light, and defocus blur conditions. Updating the neighborhood predictor on the test image suppresses spurious residuals in normal backgrounds (yellow boxes) while sharpening anomaly responses within ground-truth contours (cyan boxes).}
  \label{fig:qualitative}
\end{figure*}

\vspace{0.4em}
\noindent\textbf{Conflict coordination stabilizes TTT updates.}
The binding update alone surges VisA$\rightarrow$MVTec pixel AP from 15.2 to 43.0 ($\uparrow$27.8), while adding gradient conflict coordination reaches 44.2 and 28.3 pixel AP across both transfer directions (Table~\ref{tab:update_ablation}). Because anomalous regions yield large, erratic gradients during unsupervised test-time updates, projecting out conflicting gradient components prevents defects from corrupting the shared predictor.

\begin{table}[!htbp]
\centering
\caption{Update-rule ablation. Each entry reports
(AUROC, F1, AP). \textit{No update}
retains the static neighborhood predictor but disables
parameter adaptation during both training and inference.}
\label{tab:update_ablation}

\footnotesize 
\renewcommand{\arraystretch}{1.3} 
\setlength{\tabcolsep}{1.0mm} 

\begin{tabular}{l c c} 
\toprule
\textbf{Update rule} & \textbf{Pixel-level} & \textbf{Image-level} \\
\midrule
\multicolumn{3}{c}{\textbf{VisA$\rightarrow$MVTec}} \\
\midrule
No update          & (85.6, 23.0, 15.2) & (86.2, 89.5, 93.0) \\
Binding            & (90.9, 44.6{\tiny\textcolor{red!70!black}{$\uparrow$21.6}}, 43.0{\tiny\textcolor{red!70!black}{$\uparrow$27.8}}) & (91.7, 93.1{\tiny\textcolor{red!70!black}{$\uparrow$3.6}}, 96.2{\tiny\textcolor{red!70!black}{$\uparrow$3.2}}) \\
\rowcolor{gray!12}
Binding + Conflict & (\textbf{91.4}, \textbf{45.7}{\tiny\textcolor{red!70!black}{\textbf{$\uparrow$22.7}}}, \textbf{44.2}{\tiny\textcolor{red!70!black}{\textbf{$\uparrow$29.0}}}) & (\textbf{93.5}, \textbf{93.3}{\tiny\textcolor{red!70!black}{\textbf{$\uparrow$3.8}}}, \textbf{97.1}{\tiny\textcolor{red!70!black}{\textbf{$\uparrow$4.1}}}) \\
\midrule
\multicolumn{3}{c}{\textbf{MVTec$\rightarrow$VisA}} \\
\midrule
No update          & (94.8, 34.1, 27.3) & (87.1, 83.7, 89.2) \\
Binding            & (95.2, 34.5{\tiny\textcolor{red!70!black}{$\uparrow$0.4}}, 27.9{\tiny\textcolor{red!70!black}{$\uparrow$0.6}}) & (87.5, 83.8{\tiny\textcolor{red!70!black}{$\uparrow$0.1}}, 89.6{\tiny\textcolor{red!70!black}{$\uparrow$0.4}}) \\
\rowcolor{gray!12}
Binding + Conflict & (\textbf{95.5}, \textbf{34.7}{\tiny\textcolor{red!70!black}{\textbf{$\uparrow$0.6}}}, \textbf{28.3}{\tiny\textcolor{red!70!black}{\textbf{$\uparrow$1.0}}}) & (\textbf{88.4}, \textbf{84.5}{\tiny\textcolor{red!70!black}{\textbf{$\uparrow$0.8}}}, \textbf{90.4}{\tiny\textcolor{red!70!black}{\textbf{$\uparrow$1.2}}}) \\
\bottomrule
\end{tabular}
\end{table}

\vspace{0.4em}
\noindent\textbf{Spatial context outperforms global statistics.}
Preserving exact neighborhood-to-center correspondence (\textit{Original}) achieves the best pixel AP of 44.2 and 28.3, surpassing spatial shuffling (\textit{Shuffled}, 39.8/27.9) and channel statistics (\textit{Mean-only}, 41.5/27.7) (Table~\ref{tab:update_variant_ablation}). While global statistics capture coarse style shifts, fine-grained structural anomalies require modeling authentic local spatial dependencies to separate subtle defects from valid neighboring patterns.

\begin{table}[!htbp]
\caption{Update-signal ablation. Shuffled disrupts spatial correspondence, while Mean-only uses image-level channel statistics. Each entry reports (AUROC, $F_1$, AP).}

\label{tab:update_variant_ablation}

\small 
\renewcommand{\arraystretch}{1.2} 
\setlength{\tabcolsep}{3.4mm}

\begin{tabular}{l c c} 
\toprule
\textbf{Configuration} & \textbf{Pixel-level} & \textbf{Image-level} \\
\midrule
\multicolumn{3}{c}{\textbf{VisA$\rightarrow$MVTec}} \\
\midrule
No update & (85.6, 23.0, 15.2) & (86.2, 89.5, 93.0) \\
Shuffled  & (90.0, 42.0, 39.8) & (92.0, 92.5, 96.4) \\
Mean-only & (90.5, 43.5, 41.5) & (92.6, 92.8, 96.7) \\
\rowcolor{gray!12}
Original  & (\textbf{91.4}, \textbf{45.7}, \textbf{44.2}) & (\textbf{93.5}, \textbf{93.3}, \textbf{97.1}) \\
\midrule
\multicolumn{3}{c}{\textbf{MVTec$\rightarrow$VisA}} \\
\midrule
No update & (94.8, 34.1, 27.3) & (87.1, 83.7, 89.2) \\
Shuffled  & (94.9, 34.5, 27.9) & (87.4, 83.9, 89.6) \\
Mean-only & (95.5, 34.2, 27.7) & (87.3, 83.5, 89.4) \\
\rowcolor{gray!12}
Original  & (\textbf{95.5}, \textbf{34.7}, \textbf{28.3}) & (\textbf{88.4}, \textbf{84.5}, \textbf{90.4}) \\
\bottomrule
\end{tabular}
\end{table}

\vspace{0.4em}
\noindent\textbf{Tail-focused regularization is more effective.}
Selecting the highest-residual normal patches under the same 5\% budget achieves pixel APs of 44.2 and 28.3 on VisA$\rightarrow$MVTec and MVTec$\rightarrow$VisA, respectively, outperforming random selection (42.3 and 25.2) and full-normal regularization (42.8 and 26.2) (Table~\ref{tab:ntr_selection}). This supports targeting difficult normal patches over uniform regularization.

\begin{table}[!htbp]
    \centering
    \caption{Normal-patch selection ablation (\%). Random and tail selection use 5\% of normal patches. Metrics follow Table~\ref{tab:update_ablation}.}
    \label{tab:ntr_selection}

    \small 
    \renewcommand{\arraystretch}{1.15} 
    \setlength{\tabcolsep}{2.2mm}

    \begin{tabular}{l c c} 
        \toprule
        \textbf{Normal selection} & \textbf{Pixel-level} & \textbf{Image-level} \\
        \midrule
        \multicolumn{3}{c}{\textbf{VisA$\rightarrow$MVTec}} \\
        \midrule
        None                 & (90.5, 44.3, 42.1) & (92.3, 93.0, 96.5) \\
        All normal           & (90.8, 44.9, 42.8) & (\textbf{93.5}, 93.1, 97.0) \\
        Random 5\%           & (90.7, 44.4, 42.3) & (93.0, 93.0, 96.7) \\
        \rowcolor{gray!12}
        Highest-residual 5\% & (\textbf{91.4}, \textbf{45.7}, \textbf{44.2}) & (\textbf{93.5}, \textbf{93.3}, \textbf{97.1}) \\
        \midrule
        \multicolumn{3}{c}{\textbf{MVTec$\rightarrow$VisA}} \\
        \midrule
        None                 & (95.1, 31.9, 24.7) & (85.2, 82.1, 87.9) \\
        All normal           & (95.3, 32.9, 26.2) & (87.3, 84.0, 89.4) \\
        Random 5\%           & (95.2, 32.4, 25.2) & (85.4, 82.1, 88.0) \\
        \rowcolor{gray!12}
        Highest-residual 5\% & (\textbf{95.5}, \textbf{34.7}, \textbf{28.3}) & (\textbf{88.4}, \textbf{84.5}, \textbf{90.4}) \\
        \bottomrule
    \end{tabular}
\end{table}

\vspace{0.5em}
\noindent\textbf{Computational Efficiency.}
Table~\ref{tab:efficiency} compares the computational overhead of representative ZSAD methods. ReTTT-AD achieves the lowest burden across all metrics, requiring only 0.665M additional parameters, 1785.9 MiB peak memory, and 1.683 extra GFLOPs. Compared to VisualAD, it slashes additional parameters by 97.9\% and peak memory by 11.7\%, demonstrating the extreme efficiency of our lightweight relational adaptation.

\begin{table}[t]
\centering
\caption{Efficiency comparison on an NVIDIA RTX 3090 (batch size 1). Extra GFLOPs are relative to the CLIP image encoder.}
\label{tab:efficiency}
\small
\setlength{\tabcolsep}{3.5pt}
\renewcommand{\arraystretch}{1.15}
\begin{tabular}{lccc}
\toprule
\textbf{Method}
& \shortstack{\textbf{Extra Params}\\\textbf{(M)} $\downarrow$}
& \shortstack{\textbf{Peak Memory}\\\textbf{(MiB)} $\downarrow$}
& \shortstack{\textbf{Extra}\\\textbf{GFLOPs} $\downarrow$} \\
\midrule
AnomalyCLIP & 5.555 & 2279.5 & 213.392 \\
FAPrompt    & 9.612 & 2109.8 & 501.831 \\
AdaptCLIP     & 1.763 & 2240.1	 & 214.178 \\
VisualAD    & 31.521 & 2021.7 & 71.266 \\
\rowcolor{gray!12}
\textbf{Ours} & \textbf{0.665} & \textbf{1785.9} & \textbf{1.683} \\
\bottomrule
\end{tabular}
\end{table}

\begin{figure}[t]
    \centering
    \includegraphics[width=\linewidth]{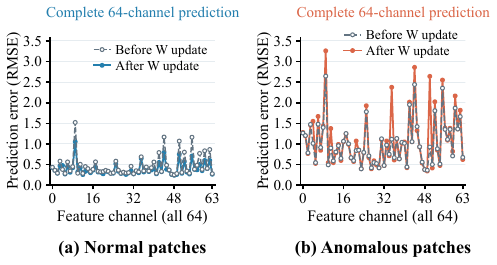}
    \caption{\textbf{Channel-wise prediction errors before and after adaptation.} Class-balanced RMSE per channel across all MVTec-AD categories for normal (left) and anomalous (right) patches. Dashed gray curves denote pre-update baselines; solid blue and orange curves show post-update errors.} 
    \label{fig:channel}
\end{figure}


\begin{figure}[t]
    \centering
    \begin{subfigure}[b]{\linewidth}
        \centering
        \includegraphics[width=1\linewidth]{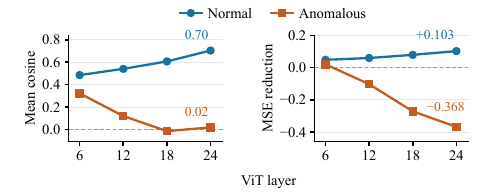} 
        \caption{Layer-wise prediction correction}
        \label{fig:residual_a}
    \end{subfigure}
    
    \vspace{0.6em} 
    
    \begin{subfigure}[b]{\linewidth}
        \centering
        \includegraphics[width=0.9\linewidth]{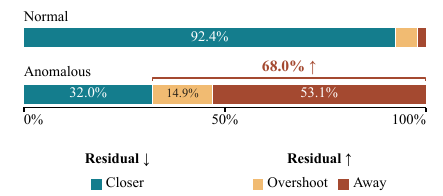} 
        \caption{Patch-level residual changes}
        \label{fig:residual_b}
    \end{subfigure}
    
    \caption{\textbf{Mechanism of differential prediction correction.}
    (a) Mean cosine similarity between prediction corrections and pre-update error directions, and mean MSE reduction across four ViT layers. Positive MSE reduction indicates lower post-update error.
    (b) Class-balanced proportions of normal and anomalous patches whose predictions move \textit{closer} to the target, \textit{overshoot} it enough to increase error, or move \textit{away} from it.}
    \label{fig:residual}
\end{figure}

\subsection{Visualization}
We analyze how test-time adaptation alters neighborhood predictions and enhances residual contrast across domains, corruptions, and feature channels. We further examine these effects across normal and anomalous patches.

\vspace{0.4em}
\noindent\textbf{TTT resists domain and corruption shifts.}
Enabling TTT suppresses background false positives (yellow boxes) while sharpening defect localization (cyan boxes) across industrial and medical benchmarks under clean, Gaussian noise, low-light, and defocus-blur conditions (Fig.~\ref{fig:qualitative}). Because visual corruptions shift global feature distributions uniformly across an image, adapting the relational predictor online can effectively absorb these style perturbations without erasing local structural discontinuities.

\vspace{0.4em}
\noindent\textbf{Adaptation suppresses normal errors across channels.}
Updating $W$ lowers prediction errors across prominent channels for normal patches while retaining or amplifying high responses on anomalous patches (Fig.~\ref{fig:channel}). This divergence indicates that dominant normal patterns drive the shared linear update, whereas out-of-distribution defects remain unmodeled by the adapted subspace.

\vspace{0.4em}
\noindent\textbf{Adaptation preferentially corrects normal predictions.}
Across the four ViT layers, prediction corrections align more closely with the pre-update error directions of normal patches and yield larger mean MSE reductions (Fig.~\ref{fig:residual}(a)). The mean cosine similarity is $0.70$ for normal patches and $0.02$ for anomalous patches, and their mean MSE decreases by $0.103$ and increases by $0.368$, respectively. At the patch level, $92.4\%$ of normal patches show lower residuals after adaptation (Fig.~\ref{fig:residual}(b)). Among anomalous patches, $53.1\%$ move away from the target and $14.9\%$ overshoot it enough to increase the residual. Together, these outcomes show that directional mismatch accounts for most of the observed residual increases in anomalous patches.

\section{Conclusion}
\label{sec:conclusion}

We presented ReTTT-AD, a zero-shot anomaly detection framework that formulates local anomaly discrimination as image-conditioned neighborhood prediction. RL-TTT adapts a lightweight spatially shared predictor to each test image through a single test-time update, while NTR suppresses spurious residuals in complex normal regions and GAR provides complementary global evidence. Extensive experiments across 14 industrial and medical benchmarks demonstrate strong cross-domain generalization and anomaly localization. Further analyses reveal that test-time adaptation preferentially reduces normal prediction errors while retaining or increasing anomalous residuals. Future work will investigate this differential correction mechanism across diverse imaging modalities.

{
  \small
  \bibliographystyle{ieeenat_fullname}
  \bibliography{main}

@String(CVPR= {IEEE Conf. Comput. Vis. Pattern Recog.})

@String(ICCV= {Int. Conf. Comput. Vis.})

@String(ECCV= {Eur. Conf. Comput. Vis.})

@String(ICLR = {Int. Conf. Learn. Represent.})

@String(AAAI = {AAAI})

@String(CVPR  = {CVPR})

@String(ICCV  = {ICCV})

@String(ECCV  = {ECCV})

@String(ICLR  = {ICLR})

@inproceedings{CLIP,
  author       = {Alec Radford and
                  Jong Wook Kim and
                  Chris Hallacy and
                  Aditya Ramesh and
                  Gabriel Goh and
                  Sandhini Agarwal and
                  Girish Sastry and
                  Amanda Askell and
                  Pamela Mishkin and
                  Jack Clark and
                  Gretchen Krueger and
                  Ilya Sutskever},
  title        = {Learning Transferable Visual Models From Natural Language Supervision},
  booktitle    = {Proceedings of the 38th International Conference on Machine Learning, {ICML}},
  pages        = {8748--8763},
  year         = {2021},
}

@inproceedings{AnomalyCLIP,
    author={Qihang Zhou and Guansong Pang and Yu Tian and Shibo He and Jiming Chen},
    title={AnomalyCLIP: Object-agnostic Prompt Learning for Zero-shot Anomaly Detection},
    booktitle={The Twelfth International Conference on Learning Representations (ICLR)},
    pages={1-33},
    year={2024},
}

@inproceedings{WinCLIP,
  author       = {Jongheon Jeong and
                  Yang Zou and
                  Taewan Kim and
                  Dongqing Zhang and
                  Avinash Ravichandran and
                  Onkar Dabeer},
  title        = {WinCLIP: Zero-/Few-Shot Anomaly Classification and Segmentation},
  booktitle    = {{IEEE/CVF} Conference on Computer Vision and Pattern Recognition,
                  {CVPR}},
  pages        = {19606--19616},
  publisher    = {{IEEE}},
  year         = {2023},
}

@inproceedings{MVTec,  
   title={MVTec AD --- A Comprehensive Real-World Dataset for Unsupervised Anomaly Detection}, 
   booktitle={2019 IEEE/CVF Conference on Computer Vision and Pattern Recognition (CVPR)}, 
   author={Bergmann, Paul and Fauser, Michael and Sattlegger, David and Steger, Carsten}, 
   year={2019}, 
}

@inproceedings{visa,
  author       = {Yang Zou and
                  Jongheon Jeong and
                  Latha Pemula and
                  Dongqing Zhang and
                  Onkar Dabeer},
  title        = {SPot-the-Difference Self-supervised Pre-training for Anomaly Detection
                  and Segmentation},
  booktitle    = {Computer Vision - {ECCV} 2022 - 17th European Conference, Tel Aviv},
  volume       = {13690},
  pages        = {392--408},
  year         = {2022},
}

@inproceedings{AdaCLIP,
  title={Adaclip: Adapting clip with hybrid learnable prompts for zero-shot anomaly detection},
  author={Cao, Yunkang and Zhang, Jiangning and Frittoli, Luca and Cheng, Yuqi and Shen, Weiming and Boracchi, Giacomo},
  booktitle={European conference on computer vision},
  pages={55--72},
  year={2024},
  organization={Springer}
}

@inproceedings{BTAD,
  author       = {Pankaj Mishra and
                  Riccardo Verk and
                  Daniele Fornasier and
                  Claudio Piciarelli and
                  Gian Luca Foresti},
  title        = {{VT-ADL:} {A} Vision Transformer Network for Image Anomaly Detection
                  and Localization},
  booktitle    = {30th {IEEE} International Symposium on Industrial Electronics, {ISIE}
                  2021, Kyoto, Japan, June 20-23, 2021},
  pages        = {1--6},
  year         = {2021},
}

@article{KSDD2,
  title={Mixed supervision for surface-defect detection: From weakly to fully supervised learning},
  author={Bo{\v{z}}i{\v{c}}, Jakob and Tabernik, Domen and Sko{\v{c}}aj, Danijel},
  journal={Computers in Industry},
  volume={129},
  pages={103459},
  year={2021},
  doi={10.1016/j.compind.2021.103459},
  publisher={Elsevier}
}

@misc{DAGM,
  title={Weakly Supervised Learning for Industrial Optical Inspection},
  author={Wieler, Matthias and Hahn, Tobias and Hamprecht, Fred A.},
  year={2007},
  howpublished={DAGM 2007 Dataset}
}

@inproceedings{selfsupervised5,
  author       = {Nicolae{-}Catalin Ristea and
                  Neelu Madan and
                  Radu Tudor Ionescu and
                  Kamal Nasrollahi and
                  Fahad Shahbaz Khan and
                  Thomas B. Moeslund and
                  Mubarak Shah},
  title        = {Self-Supervised Predictive Convolutional Attentive Block for Anomaly
                  Detection},
  booktitle    = {{IEEE/CVF} Conference on Computer Vision and Pattern Recognition,
                  {CVPR} 2022},
  pages        = {13566--13576},
  year         = {2022},
}

@inproceedings{MFAD,
  title={Adapting Visual-Language Models for Generalizable Anomaly Detection in Medical Images},
  author={Huang, Chaoqin and Jiang, Aofan and Feng, Jinghao and Zhang, Ya and Wang, Xinchao and Wang, Yanfeng},
  booktitle={IEEE/CVF Conference on Computer Vision and Pattern Recognition (CVPR)},
  year={2024},
}

@article{DINOv2,
  title={{DINOv2}: Learning robust visual features without supervision},
  author={Oquab, Maxime and Darcet, Timoth{\'e}e and Moutakanni, Th{\'e}o and Vo, Huy and Szafraniec, Marc and Khalidov, Vasil and Fernandez, Pierre and Haziza, Daniel and Massa, Francisco and El-Nouby, Alaaeldin and others},
  journal={Transactions on Machine Learning Research},
  year={2024}
}

@inproceedings{DTD-Synthetic,
  title={Zero-shot versus many-shot: Unsupervised texture anomaly detection},
  author={Aota, Toshimichi and Tong, Lloyd Teh Tzer and Okatani, Takayuki},
  booktitle={Proceedings of the IEEE/CVF Winter Conference on Applications of Computer Vision},
  pages={5564--5572},
  year={2023}
}

@inproceedings{VCP,
  title={VCP-CLIP: A visual context prompting model for zero-shot anomaly segmentation},
  author={Zhen Qu and Xian Tao and Mukesh Prasad and Fei Shen and Zhengtao Zhang and Xinyi Gong and Guiguang Ding},
  booktitle={European Conference on Computer Vision},
  year={2024}
}

@inproceedings{PaDiM,
  title={Padim: a patch distribution modeling framework for anomaly detection and localization},
  author={Defard, Thomas and Setkov, Aleksandr and Loesch, Angelique and Audigier, Romaric},
  booktitle={International conference on pattern recognition},
  pages={475--489},
  year={2021},
  organization={Springer}
}

@inproceedings{PatchCore,
  title={Towards total recall in industrial anomaly detection},
  author={Roth, Karsten and Pemula, Latha and Zepeda, Joaquin and Sch{\"o}lkopf, Bernhard and Brox, Thomas and Gehler, Peter},
  booktitle={2022 IEEE/CVF Conference on Computer Vision and Pattern Recognition (CVPR)},
  pages={14298--14308},
  year={2022},
  organization={IEEE}
}

@article{CutPaste,
  title={CutPaste: Self-Supervised Learning for Anomaly Detection and Localization},
  author={Chun-Liang Li and Kihyuk Sohn and Jinsung Yoon and Tomas Pfister},
  journal={2021 IEEE/CVF Conference on Computer Vision and Pattern Recognition (CVPR)},
  year={2021},
  pages={9659-9669}
}

@inproceedings{AdaptCLIP,
  title={AdaptCLIP: Adapting CLIP for Universal Visual Anomaly Detection},
  author={Gao, Bin-Bin and Zhou, Yue and Yan, Jiangtao and Cai, Yuezhi and Zhang, Weixi and Wang, Meng and Liu, Jun and Liu, Yong and Wang, Lei and Wang, Chengjie},
  booktitle={AAAI},
  year={2026}
}

@article{Auto2,
  title={Memorizing Normality to Detect Anomaly: Memory-Augmented Deep Autoencoder for Unsupervised Anomaly Detection},
  author={Dong Gong and Lingqiao Liu and Vuong Le and Budhaditya Saha and Moussa Reda Mansour and Svetha Venkatesh and Anton van den Hengel},
  journal={2019 IEEE/CVF International Conference on Computer Vision (ICCV)},
  year={2019},
  pages={1705-1714}
}

@article{OCT17,
  title={Identifying medical diagnoses and treatable diseases by image-based deep learning},
  author={Kermany, Daniel S and Goldbaum, Michael and Cai, Wenjia and Valentim, Carolina CS and Liang, Huiying and Baxter, Sally L and McKeown, Alex and Yang, Ge and Wu, Xiaokang and Yan, Fangbing and others},
  journal={Cell},
  volume={172},
  number={5},
  pages={1122--1131},
  year={2018},
  publisher={Elsevier}
}

@inproceedings{BrainMRI,
  title={Multiresolution knowledge distillation for anomaly detection},
  author={Salehi, Mohammadreza and Sadjadi, Niousha and Baselizadeh, Soroosh and Rohban, Mohammad H and Rabiee, Hamid R},
  booktitle={Proceedings of the IEEE/CVF conference on computer vision and pattern recognition},
  pages={14902--14912},
  year={2021}
}

@article{HIS,
  title={Diagnostic assessment of deep learning algorithms for detection of lymph node metastases in women with breast cancer},
  author={Ehteshami Bejnordi, Babak and Veta, Mitko and Johannes van Diest, Paul and van Ginneken, Bram and Karssemeijer, Nico and Litjens, Geert and van der Laak, Jeroen A. W. M. and Hermsen, Meyke and Manson, Quirine F. and others},
  journal={{JAMA}},
  volume={318},
  number={22},
  pages={2199--2210},
  year={2017},
  doi={10.1001/jama.2017.14585}
}

@article{CVC-ClinicDB,
  title={WM-DOVA maps for accurate polyp highlighting in colonoscopy: Validation vs. saliency maps from physicians},
  author={Jorge Bernal and Francisco Javier S{\'a}nchez and Gl{\`o}ria Fern{\'a}ndez-Esparrach and Debora Gil and Cristina Rodr{\'i}guez de Miguel and Fernando Vilari{\~n}o},
  journal={Computerized medical imaging and graphics : the official journal of the Computerized Medical Imaging Society},
  year={2015},
  volume={43},
  pages={
          99-111
        }
}

@inproceedings{Endo,
  title={The EndoTect 2020 challenge: evaluation and comparison of classification, segmentation and inference time for endoscopy},
  author={Hicks, Steven A and Jha, Debesh and Thambawita, Vajira and Halvorsen, P{\aa}l and Hammer, Hugo L and Riegler, Michael A},
  booktitle={International Conference on Pattern Recognition},
  pages={263--274},
  year={2021},
  organization={Springer}
}

@inproceedings{Kvasir,
  title={{Kvasir-SEG}: A Segmented Polyp Dataset},
  author={Jha, Debesh and Smedsrud, Pia H. and Riegler, Michael A. and Halvorsen, P{\aa}l and {de Lange}, Thomas and Johansen, Dag and Johansen, H{\aa}vard D.},
  booktitle={MultiMedia Modeling},
  series={Lecture Notes in Computer Science},
  volume={11962},
  pages={451--462},
  year={2020},
  publisher={Springer},
  doi={10.1007/978-3-030-37734-2_37}
}

@inproceedings{VisualAD,
  author={Hou, Yanning and Li, Peiyuan and Liu, Zirui and Wang, Yitong and Ruan, Yanran and Qiu, Jianfeng and Xu, Ke},
  title={VisualAD: Language-Free Zero-Shot Anomaly Detection via Vision Transformer},
  booktitle={Proceedings of the IEEE/CVF Conference on Computer Vision and Pattern Recognition (CVPR)},
  month={June},
  year={2026},
  pages={21346--21356}
}

@article{ISIC,
  author={Gutman, David and Codella, Noel C. F. and Celebi, Emre and Helba, Brian and Marchetti, Michael and Mishra, Nabin and Halpern, Allan},
  title={Skin Lesion Analysis toward Melanoma Detection: A Challenge at the International Symposium on Biomedical Imaging ({ISBI}) 2016, hosted by the International Skin Imaging Collaboration ({ISIC})},
  journal={arXiv preprint arXiv:1605.01397},
  year={2016},
  doi={10.48550/arXiv.1605.01397},
  url={https://arxiv.org/abs/1605.01397}
}

@inproceedings{FAPrompt,
  author={Zhu, Jiawen and Ong, Yew-Soon and Shen, Chunhua and Pang, Guansong},
  title={Fine-grained Abnormality Prompt Learning for Zero-shot Anomaly Detection},
  booktitle={Proceedings of the IEEE/CVF International Conference on Computer Vision (ICCV)},
  month={October},
  year={2025},
  pages={22241--22251},
  url={https://openaccess.thecvf.com/content/ICCV2025/html/Zhu_Fine-grained_Abnormality_Prompt_Learning_for_Zero-shot_Anomaly_Detection_ICCV_2025_paper.html}
}

@article{MVTecAD2,
  author  = {Heckler-Kram, Lars and
             Neudeck, Jan-Hendrik and
             Scheler, Ulla and
             K{\"o}nig, Rebecca and
             Steger, Carsten},
  title   = {The MVTec AD 2 Dataset: Advanced Scenarios
             for Unsupervised Anomaly Detection},
  journal = {International Journal of Computer Vision},
  volume  = {134},
  year    = {2026},
  doi     = {10.1007/s11263-026-02743-0}
}

@inproceedings{sun2020test,
  title={Test-time training with self-supervision for generalization under distribution shifts},
  author={Sun, Yu and Wang, Xiaolong and Liu, Zhuang and Miller, John and Efros, Alexei and Hardt, Moritz},
  booktitle={International conference on machine learning},
  pages={9229--9248},
  year={2020},
  organization={PMLR}
}

@article{gandelsman2022test,
  title={Test-time training with masked autoencoders},
  author={Gandelsman, Yossi and Sun, Yu and Chen, Xinlei and Efros, Alexei},
  journal={Advances in Neural Information Processing Systems},
  volume={35},
  pages={29374--29385},
  year={2022}
}

@inproceedings{DRAEM,
  title={DR{\AE}M--A discriminatively trained reconstruction embedding for surface anomaly detection},
  author={Zavrtanik, Vitjan and Kristan, Matej and Sko{\v{c}}aj, Danijel},
  booktitle={2021 IEEE/CVF International Conference on Computer Vision (ICCV)},
  pages={8310--8319},
  year={2021},
  organization={IEEE}
}

@article{UniAD,
  title={A unified model for multi-class anomaly detection},
  author={You, Zhiyuan and Cui, Lei and Shen, Yujun and Yang, Kai and Lu, Xin and Zheng, Yu and Le, Xinyi},
  journal={Advances in Neural Information Processing Systems},
  volume={35},
  pages={4571--4584},
  year={2022}
}

@inproceedings{EfficientAD,
  title={Efficientad: Accurate visual anomaly detection at millisecond-level latencies},
  author={Batzner, Kilian and Heckler, Lars and K{\"o}nig, Rebecca},
  booktitle={2024 IEEE/CVF Winter Conference on Applications of Computer Vision (WACV)},
  pages={127--137},
  year={2024},
  organization={IEEE}
}

@article{tpt,
  title={Test-time prompt tuning for zero-shot generalization in vision-language models},
  author={Shu, Manli and Nie, Weili and Huang, De-An and Yu, Zhiding and Goldstein, Tom and Anandkumar, Anima and Xiao, Chaowei},
  journal={Advances in Neural Information Processing Systems},
  volume={35},
  pages={14274--14289},
  year={2022}
}

@inproceedings{liu2023simplenet,
  title={Simplenet: A simple network for image anomaly detection and localization},
  author={Liu, Zhikang and Zhou, Yiming and Xu, Yuansheng and Wang, Zilei},
  booktitle={2023 IEEE/CVF Conference on Computer Vision and Pattern Recognition (CVPR)},
  pages={20402--20411},
  year={2023},
  organization={IEEE}
}

@inproceedings{deng2022anomaly,
  title={Anomaly detection via reverse distillation from one-class embedding},
  author={Deng, Hanqiu and Li, Xingyu},
  booktitle={2022 IEEE/CVF conference on computer vision and pattern recognition (CVPR)},
  pages={9727--9736},
  year={2022},
  organization={IEEE}
}

@article{liu2021ttt++,
  title={Ttt++: When does self-supervised test-time training fail or thrive?},
  author={Liu, Yuejiang and Kothari, Parth and Van Delft, Bastien and Bellot-Gurlet, Baptiste and Mordan, Taylor and Alahi, Alexandre},
  journal={Advances in Neural Information Processing Systems},
  volume={34},
  pages={21808--21820},
  year={2021}
}

@inproceedings{gudovskiy2022cflow,
  title={Cflow-ad: Real-time unsupervised anomaly detection with localization via conditional normalizing flows},
  author={Gudovskiy, Denis and Ishizaka, Shun and Kozuka, Kazuki},
  booktitle={2022 IEEE/CVF winter conference on applications of computer vision (WACV)},
  pages={1819--1828},
  year={2022},
  organization={IEEE}
}
}
\end{document}